%% file: main.tex
\documentclass[conference]{IEEEtran}
\IEEEoverridecommandlockouts
\usepackage{cite}
\usepackage{amsmath,amssymb,amsfonts}
\usepackage{algorithmic}
\usepackage{graphicx}
\usepackage{amssymb}
\usepackage{textcomp}
\usepackage{xcolor}
\def\BibTeX{{\rm B\kern-.05em{\sc i\kern-.025em b}\kern-.08em
    T\kern-.1667em\lower.7ex\hbox{E}\kern-.125emX}}

\usepackage{amsmath,amsfonts}
\usepackage{algorithmic}
\usepackage{algorithm}
\usepackage{array}
\usepackage[caption=false,font=normalsize,labelfont=sf,textfont=sf]{subfig}
\usepackage{textcomp}
\usepackage{stfloats}
\usepackage{url}
\usepackage{verbatim}
\usepackage{graphicx}
\usepackage[utf8]{inputenc}
\usepackage{graphicx}
\usepackage{booktabs}
\usepackage{multirow}
\usepackage{siunitx}
\usepackage{cite}
\usepackage{multirow}
\usepackage{graphicx}
\usepackage{caption}
\usepackage{booktabs}
\usepackage{rotating}
\usepackage{array}
\usepackage{siunitx}
\usepackage{float}
\usepackage{overpic}
\usepackage{subcaption}
\usepackage{subfig}
\usepackage{pgfplots}
\usepackage{siunitx}
\usepackage{booktabs} 
\usepackage{multirow}
\usepackage{graphicx}
\usepackage{caption}
\usepackage{color}

\usepackage{rotating}
\usepackage{array}
\usepackage{siunitx}
\usepackage{float}
\usepackage{overpic}
\usepackage{subcaption}
\usepackage{pgfplots}
\usepackage{hyperref}
\begin{document}

\title{From Part to Whole: 3D Generative World Model with an  Adaptive Structural Hierarchy}







\author{
\IEEEauthorblockN{Bi'an Du*}
\IEEEauthorblockA{
Wangxuan Institute of \\Computer Technology,\\ Peking University\\
Beijing, China\\
pkudba@stu.pku.edu.cn
}
\and
\IEEEauthorblockN{Daizong Liu*}
\IEEEauthorblockA{
Institute for Math \& AI,\\ Wuhan University\\
Wuhan, China\\
daizongliu@whu.edu.cn
}
\and
\IEEEauthorblockN{Pufan Li}
\IEEEauthorblockA{
Wangxuan Institute of \\Computer Technology, \\ Peking University\\
Beijing, China\\
lipufan@pku.edu.cn
}
\and
\IEEEauthorblockN{Wei Hu$^\dagger$\\}
\IEEEauthorblockA{
Wangxuan Institute of \\Computer Technology, \\ Peking University\\
Beijing, China\\
forhuwei@pku.edu.cn
}
\thanks{* Equal contribution. \protect\\ \hspace*{0.7em} $^\dagger$ Corresponding author: Wei Hu (forhuwei@pku.edu.cn).}

}

\maketitle

\begin{abstract}
Single-image 3D generation lies at the core of vision-to-graphics models in the real world. 
However, it remains a fundamental challenge to achieve reliable generalization across diverse semantic categories and highly variable structural complexity under sparse supervision.
Existing approaches typically model objects in a monolithic manner or rely on a fixed number of parts, including recent part-aware models such as PartCrafter, which still require a labor-intensive user-specified part count. Such designs easily lead to overfitting, fragmented or missing structural components, and limited compositional generalization when encountering novel object layouts.
To this end, this paper rethinks single-image 3D generation as learning an adaptive part–whole hierarchy in the flexible 3D latent space. 
We present a novel part-to-whole 3D generative world model that autonomously discovers latent structural slots by inferring soft and compositional masks directly from image tokens.
Specifically, an adaptive slot-gating mechanism dynamically determines the slot-wise activation probabilities and smoothly consolidates redundant slots within different objects, ensuring that the emergent structure remains compact yet expressive across categories. 
Each distilled slot is then aligned to a learnable, class-agnostic prototype bank, enabling powerful cross-category shape sharing and denoising through universal geometric prototypes in the real world.
Furthermore, a lightweight 3D denoiser is introduced to reconstruct geometry and appearance via unified diffusion objectives. Experiments show consistent gains in cross-category transfer and part-count extrapolation, and ablations confirm complementary benefits of the prototype bank for shape-prior sharing as well as slot-gating for structural adaptation.   
\end{abstract}


\input{sec/1_intro}

\input{sec/2_related}

\input{sec/3_modeling}
\input{sec/4_experiment}
\input{sec/5_conclusion}

\bibliographystyle{IEEEbib}
\bibliography{main}

\end{document}

%% file: sec/1_intro.tex
\section{Introduction}
\label{sec:intro}

Single-image 3D generation \cite{fahim2021single,hong2023lrm} is central to various real-world multimedia and vision-to-graphics applications, such as AR/VR content creation \cite{hong2023lrm,zhang2025behave}, virtual retrieval \cite{liu2021context}, digital modification \cite{liu2025seeing}, and interactive 3D editing \cite{liu2024survey}. Given a single RGB image, the goal of this task is to recover a plausible 3D object or scene that is geometrically consistent, visually faithful, and structurally coherent. Recent advances \cite{yu2021pixelnerf,liu2023robust,liu2022imperceptible,du2021self,chen2022deep,du2024generative,du2024multi} in implicit neural representations, radiance fields, Gaussian splatting, and latent diffusion or flow models have significantly improved 3D generation from sparse views. Yet in the single-image setting, the problem remains highly ill-posed, where only a single view is provided for many valid 3D explanations. 
Also, robustness under large shape variation, occlusion and complex layouts is still far from solved.

\begin{figure}
\centering
\includegraphics[width=\columnwidth]{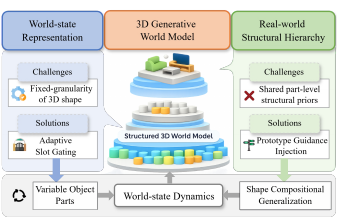}
\caption{We rethink single-image 3D generation as learning a structured 3D world state with an adaptive part–whole hierarchy. Specifically, latent structural slots are dynamically gated per instance, and a class-agnostic prototype bank provides shared part-level priors for improved structural coherence and generalization.}
\vspace{-3mm}
\label{fig:teaser}
\end{figure}

Although several approaches \cite{lin2025partcrafter,li2025triposg,tochilkin2024triposr,li2025geometry} have been proposed to handle single-image 3D generation in recent years, as shown in Fig.~\ref{fig:teaser}, they still face two key limitations when applied to real-world scenarios. (1) \textit{Fixed-granularity shape representations}: most methods either use a monolithic 3D representation or assume a fixed part decomposition per instance. In particular, image-to-3D generators that map an input to a single latent code often struggle to recover thin, articulated, or occluded structures, while part-based methods usually require a predefined number of parts, easily resulting in over-fragmentation on simple objects and under-segmentation on complicated realistic scenes. Even the latest compositional latent frameworks that can generate multiple parts still require labor-intensive efforts to decide how many parts to generate \cite{lin2025partcrafter} in advance. (2) \textit{Lack of shared common part-level structural priors}: previous works explicitly model the object parts by treating them as independent components without using commonsense knowledge of daily structural priors during the reconstruction process. That is, there is no mechanism to capture recurring elements such as legs, handles, shelves, or screens as reusable patterns across categories, limiting compositional generalization and making the learned representation less suitable as a structured generalized model for 3D real-world generation.


Hence, based on the above observations, this work addresses the two limitations by viewing challenging single-image 3D generation as learning an adaptive 3D structural hierarchy from part to whole in the flexible latent space. \textit{First}, to overcome the reliance on a fixed part count, we represent each object or scene with a set of 3D latent part slots and learn a gating module that predicts slot-wise activation probabilities from the input image, so the model can activate a variable number of parts per instance. This is further combined with a masked flow-matching objective that supervises only activated slots corresponding to ground-truth parts, preserving the reconstruction ability of a 3D Diffusion Transformer (DiT) backbone while encouraging stable and image-driven structural decompositions. 
\textit{Second}, to overcome the missing shared structural priors, we introduce a class-agnostic prototype bank in the latent slot space during the reconstruction process. Each active slot is softly aligned to a combination of global geometric prototypes, and prototype-guided residuals are injected back into the latents for promoting cross-category sharing of recurring 3D part patterns without imposing a hard bottleneck on fine-grained shape expressiveness. Together, these components realize a three-level hierarchy, from whole objects or scenes, through adaptive structural slots, down to detailed prototypes and tokens, which move towards a 3D generative world model with structured and reusable building blocks for diverse objects.

In sum, our main contributions are as follows:
\begin{itemize}
    \item In this paper, we present a new part-to-whole 3D world model, formulating single-image 3D generation as learning an adaptive structural hierarchy, where each object or scene is represented by latent structural slots.
    \item We propose an adaptive slot-gating mechanism to learn variable object parts, together with a class-agnostic prototype bank that injects prototype-guided residuals into part latents, thereby achieving shape compositional generalization while preserving fine-grained details.
    \item We evaluate our model on both object- and scene-level benchmarks, showing improvement on cross-category generalization and robustness. Ablations confirm the benefits of adaptive slot-gating and the prototype bank for structural adaptation and geometric prior sharing.
\end{itemize}


%% file: sec/2_related.tex
\section{Related Work}
\label{sec:related}

\subsection{Part-level Shape Generation}
Part-level 3D generation aims to produce decomposable assets with editable components, but existing solutions differ in how they obtain parts and how they handle varying part counts. PartCrafter \cite{lin2025partcrafter} is a representative unified model that directly denoises multiple part latents from a single image without external segmentation, and can hallucinate plausible amodal parts beyond the visible regions. However, it still assumes a user-specified part number at inference time. PartPacker \cite{tang2025efficient} tackles the unknown part count by generating an arbitrary number of complete parts from one image in fixed time through a dual packing representation, but its packing-driven design does not explicitly organize parts into a reusable latent prior shared across categories. X-Part \cite{yan2025x} instead targets controllable, high-fidelity part decomposition/generation with bounding-box prompts and injected point-wise semantics, focusing on promptable and editable part creation rather than image-driven adaptive part discovery. These works motivate our goal of learning an image-conditioned adaptive part-whole-hierarchy while introducing a compact prototype bank to provide reusable part-level structural priors.

\subsection{3D Scene Generation}
Scene-level generative modeling often treats scenes as collections of objects (instances) and focuses on layout, spatial relations, and scalability. ATISS synthesizes indoor scenes autoregressively as unordered sets of objects conditioned on room type and floor plan, enabling controllable and diverse layout generation \cite{paschalidou2021atiss}. For single-image scene generation, MIDI extends pretrained image-to-3D object generators to a multi-instance diffusion framework that generates multiple 3D instances jointly with learned inter-object interactions \cite{huang2025midi}. Beyond bounded indoor scenes, WorldGrow targets unbounded 3D world synthesis via hierarchical block-wise generation with block inpainting and coarse-to-fine refinement \cite{li2025worldgrow}. While these methods advance scene compositionality and scale, they typically operate at the instance/layout level rather than exposing a part-to-whole structural hierarchy with reusable part prototypes that generalize across categories and structural complexity.

%% file: sec/3_modeling.tex
\section{Method}
\label{sec:method}

\begin{figure*}
  \centering
  \vspace{-5mm}
  \includegraphics[width=\textwidth]{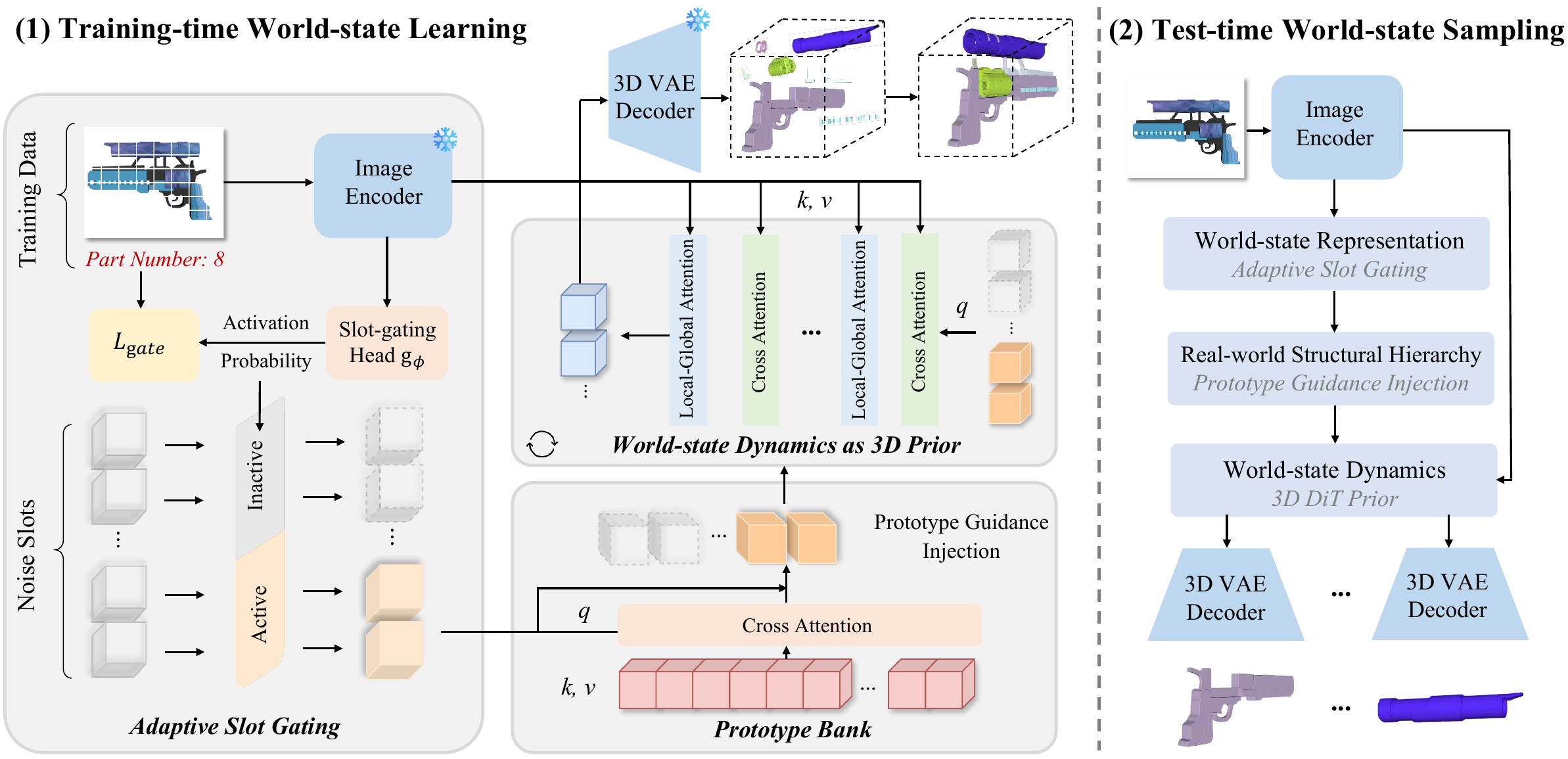}
  \caption{Framework of our part-to-whole 3D generative world model. Given a single-view RGB image, our model builds a structured 3D world state with flexible-capacity part slots. An adaptive slot-gating head selects active slots, a prototype bank injects shared part priors, and a pretrained 3D DiT rectified-flow backbone denoises slot latents with canonical-masked supervision. Decoded part meshes are finally composed into the 3D object or scene for generation.}
  \label{fig:framework}
\vspace{-2mm}
\end{figure*}

This section details the proposed single-view 3D generative world model based on adaptive structural hierarchy, as illustrated in Fig.~\ref{fig:framework}. Our design is built on a pretrained PartCrafter-style 3D VAE + DiT rectified flow backbone, and extends it with \textit{i)} latent structural slots and masked flow supervision to preserve and stabilize the pretrained geometric prior, \textit{ii)} an adaptive slot-gating head to avoid manually specifying the part count at inference, and \textit{iii)} a class-agnostic prototype bank to encourage cross-category sharing of recurring real-world part patterns.

\subsection{3D Generative World Model Backbone}
\noindent \textbf{Definition of 3D World States.}
We rethink single-view 3D generation as learning a conditional generative world model over structured 3D world states. Given a single RGB image $I$, the model predicts a plausible yet coherent world state $\mathcal{O}$ that exposes explicit part structure and supports completion under occlusion. Concretely, we represent the world state as a set of parts $\mathcal{O} = \{\mathcal{P}_i\}_{i=1}^{\hat{N}}$, where each $\mathcal{P}_i$ is a local 3D mesh or point cloud encoded by a small set of latent structural slots tied to part-level geometry, which are adaptively activated per instance. To instantiate this world model, we build on a pretrained 3D VAE + DiT rectified-flow generator, inheriting its strong category-agnostic latent 3D prior and its ability to generate plausible unseen geometry. 


\noindent \textbf{Representing World States using Structured Latent Slots.}
To expose structural hierarchy while remaining compatible with the pretrained backbone, we first define a fixed-capacity latent slot representation in the 3D VAE latent space.
Each training asset is assumed to have an object-level canonical part decomposition into $\{\mathcal{P}_i^{\text{gt}}\}_{i=1}^{N_{\text{obj}}}$, where $N_{\text{obj}}$ is the canonical number of semantic parts for this object. A pretrained 3D VAE encoder $f_{\text{enc}}$ maps each part mesh to a set of latent tokens as $\{\boldsymbol{z}_{ij}\}_{j=1}^K = f_{\text{enc}}(\mathcal{P}_i^{\text{gt}})$ and $\boldsymbol{z}_{ij} \in \mathbb{R}^C,$
where $K$ is the number of tokens per part and $C$ is the latent dimension.

We introduce a global maximum number of slots $P_{\max}$ and require the DiT backbone to always operate on a latent tensor $\boldsymbol{Z}_0 \in \mathbb{R}^{P_{\max} \times K \times C}$.
For an object with $N_{\text{obj}}$ canonical parts, we place the corresponding part latents into the top $N_{\text{obj}}$ slots and fill the remaining $P_{\max} - N_{\text{obj}}$ slots with null latents. These null latents are initialized from a learned null embedding and are detached from gradients so that they act as inert placeholders rather than trainable tokens.
We define a ground-truth slot mask $m_i$ to distinguish real-part slots and null slots:
\begin{equation}
  m_i =
  \begin{cases}
    1, & i \le N_{\text{obj}}, \\
    0, & i > N_{\text{obj}}.
  \end{cases}
\end{equation}
We also summarize each slot by an averaged embedding $\boldsymbol{s}_i = \frac{1}{K} \sum_{j=1}^K \boldsymbol{z}_{ij}$, which captures the coarse semantic and geometric information of the part associated with slot $i$.


\noindent \textbf{World-state Dynamics as Flow with Masked Supervision.}
\label{subsec:flow}
On top of the slot-wise latent representation, we adopt a rectified-flow formulation similar to \cite{lin2025partcrafter, tang2025efficient} in the 3D VAE latent space. Given a
clean latent tensor $\boldsymbol{Z}_0$ and Gaussian noise $\epsilon$ of the same shape,
we sample a time $t \sim \mathcal{U}(0,1)$ and form a noisy latent: $\boldsymbol{Z}_t = t \boldsymbol{Z}_0 + (1-t)\,\epsilon$.
The DiT backbone $v_\theta$ takes $(\boldsymbol{Z}_t, t, c)$ as input, where $c$ denotes image conditioning features extracted from the input RGB image, and predicts a velocity field $v_\theta(\boldsymbol{Z}_t, t, c) \in \mathbb{R}^{P_{\max} \times K \times C}$.

In our structured setting, only those slots that correspond to canonical parts (\textit{i.e.}, $m_i = 1$) should be supervised, while null slots should act as placeholders for flexible learning. We therefore adopt a masked flow-matching loss as:
\begin{equation}
  \mathcal{L}_{\text{m-flow}} =
  \mathbb{E}_{t, \epsilon}
  \sum_{i=1}^{P_{\max}} m_i
  \sum_{j=1}^{K}
  \big\|
    v_\theta(\boldsymbol{z}_{ij,t}, t, c)
    - (\boldsymbol{z}_{ij,0} - \epsilon_{ij})
  \big\|_2^2,
\end{equation}
where $\boldsymbol{z}_{ij,t}$ and $\boldsymbol{z}_{ij,0}$ denote the noisy and clean latents for token $j$ in slot $i$, respectively.
The loss supervises only canonical real-part slots, preserving the reconstruction ability of the pretrained 3D DiT while preventing unstable gradients from dummy slots. Each slot is decoded by the pretrained 3D VAE as $\hat{\mathcal{P}}_i = f_{\text{dec}}(\{\boldsymbol{z}_{ij}\}_{j=1}^K)$. Crucially, the mask $m_i$ in $\mathcal{L}_{\text{flow}}$ is always the ground-truth canonical mask, independent of the predicted slot gating.



\subsection{Fitting Diverse Object Parts with Adaptive Slot Gating}
\label{subsec:gating}

To overcome the reliance on a fixed part count, we introduce an adaptive slot-gating head that predicts which latent structural slots should be activated for diverse image objects. Given the feature map $f_{\text{img}}$ of input image $I$ encoded via DINOv2 \cite{oquab2023dinov2}, our slot-gating head $g_\phi$ maps $f_{\text{img}}$ to an activation probability vector $\hat{\boldsymbol{\alpha}} = g_\phi(f_{\text{img}}),$ where $\hat{\alpha}_i\in (0, 1)$ measures how likely slot $i$ should be used for this instance.
In particular, the gating target is object-level canonical structure, not view-dependent visibility. Even if a part is occluded in the image, the canonical decomposition still considers it present.

For training stability, we decouple gating from flow supervision. 
During backbone fine-tuning, the flow mask is always the ground-truth canonical mask $m_i$, not the predicted $\hat{\alpha}_i$. The gating head is trained only by gating losses, so early gating noise does not change which slots the backbone is supervised on.
We train the slot-gating head with two losses. Firstly, a slot-wise activation loss encourages each $\hat{\alpha}_i$ to match $m_i$ via binary cross-entropy, pushing canonical part slots towards high activation and null slots towards low activation:
\begin{equation}
\mathcal{L}_{\text{ce}}
=
-\frac{1}{P_{\max}}
\sum_{i=1}^{P_{\max}}
\left[
m_i \log \hat{\alpha}_i
+
\left(1-m_i\right)\log\left(1-\hat{\alpha}_i\right)
\right].
\end{equation}

Secondly, a soft count loss constrains the sum of activations to match the canonical part count:
\begin{equation}
  \mathcal{L}_{\text{count}} =
  \Big( \sum_{i=1}^{P_{\max}} \hat{\alpha}_i - N_{\text{obj}} \Big)^2.
\end{equation}

Overall, the final training objective is:
\begin{equation}
    \mathcal{L}_{\text{gate}}=\lambda_{\text{ce}} \cdot \mathcal{L}_{\text{ce}} + \lambda_{\text{count}} \cdot \mathcal{L}_{\text{count}}.   
\end{equation}

At inference time, we use $\hat{\boldsymbol{\alpha}}$ to obtain a variable number of active slots. A simple strategy is to select $\mathcal{S} = \{ i \mid \hat{\alpha}_i > \tau \}$ for a fixed threshold $\tau$. Only slots in $\mathcal{S}$ are then used for later sampling and decoding, yielding an adaptive part decomposition per instance.

\subsection{Real-world Structural Hierarchy via Prototype Bank}
\label{subsec:prototype}

Although adaptive gating mechanisms address the limitations of fixed component counts, relying solely on slots can still isolate components from different objects and hinder cross-category reuse. To address this, we further propose a real-world prototype bank to learn shared component-level geometric priors in the slot embedding space, thereby improving compositional generalization capabilities.
In particular, we maintain $M$ learnable prototype vectors $\boldsymbol{P} = \{ \boldsymbol{p}_k \}_{k=1}^M$ with $M$ chosen to be much smaller than the total number of distinct part shapes in the training set. For each slot embedding $\boldsymbol{s}_i$, we compute similarity to all prototypes to obtain the soft assignment and the resulting prototype-aligned embedding:
\begin{equation}
  \boldsymbol{w}_{ik} =
  \frac{
    \exp\big( \boldsymbol{s}_i^\top \boldsymbol{p}_k / \sqrt{C} \big)
  }{
    \sum_{k\prime} \exp\big( \boldsymbol{s}_i^\top \boldsymbol{p}_{k\prime} / \sqrt{C} \big)
  }, \quad \tilde{\boldsymbol{s}}_i = \sum_{k=1}^M \boldsymbol{w}_{ik} \boldsymbol{p}_k,
\end{equation}
which can be interpreted as reconstructing $\boldsymbol{s}_i$ from a small set of global geometric prototypes.

We inject prototype guidance back into the part latents via a residual update. For slot $i$ and its tokens $\boldsymbol{z}_{ij}$, we apply $\boldsymbol{z}_{ij} = z_{ij} + \beta \,\tilde{\boldsymbol{s}}_i,$
where $\beta$ is a small scalar controlling the strength of prototype guidance. Because the update is residual and small, the prototype bank acts as a soft geometric prior rather than a hard bottleneck, encouraging slots with similar geometry to cluster in the prototype space, while preserving the backbone's capacity for fine-grained detail.

To ensure that prototypes are both reusable and truly useful for generation, we learn a prototype bank by combining structural regularization and generation signals. Firstly, we align the slot embedding and prototype hybrid embedding on the real slots using the reconstruction loss $\mathcal{L}_{\text{rec}}=\sum_{i=1}^Nm_i\|\boldsymbol{s}_i-\tilde{\boldsymbol{s}}_i \|_2^2$. Secondly, to avoid overly uniform distribution, we add a negative entropy regularization $\mathcal{L}_{\text{ent}}=\sum_{i}m_i \sum_k \alpha_{ik} \log \alpha_{ik}$ to encourage each slot to primarily use a few prototypes. Combining the masked flow loss calculated with backbone flow matching $\mathcal{L}_{\text{m-flow}}$, the overall prototype objective is:
\begin{equation}
    \mathcal{L}_{\text{all}}=\mathcal{L}_{\text{rec}}+\lambda_{\text{ent}} \cdot \mathcal{L}_{\text{ent}}+\lambda_{\text{flow}} \cdot \mathcal{L}_{\text{m-flow}}.
\end{equation}

\subsection{Overall Training of Structured 3D World State}
\label{subsec:training_inference}

 
\textit{Step 1: Adaptive Slot and Prototype Warm-up.} We freeze the 3D VAE and most DiT layers, and train only the slot-gating head and the prototype bank.
This stage focuses on learning to infer canonical slot activations from images and to organize slot embeddings into prototypes, without perturbing the core 3D prior.

\textit{Step 2: Backbone-aware Joint Fine-tuning.} We unfreeze selected DiT blocks and jointly optimize the masked flow loss, slot-gating losses, prototype losses, and optional geometry losses with carefully tuned weights. The masked flow objective keeps the backbone close to its original reconstruction behavior, while the structural and prototype modules gradually reshape the latent space into an adaptive, prototype-aware structural hierarchy.

\textit{Step 3: Test-Time World-state Sampling.} At test time, given a single RGB image, we first encode it to obtain $f_{\text{img}}$ and apply the slot-gating head to obtain $\hat{\boldsymbol{\alpha}}$. We select an adaptive set of active slots $\mathcal{S}$ based on $\hat{\boldsymbol{\alpha}}$, initialize latent
tokens for these slots from the prior, and run the rectified-flow sampling process through the DiT backbone with prototype-guided residuals applied as during training. Finally, we decode each active slot with the 3D VAE decoder to obtain local part geometries $\hat{\mathcal{P}}_i$, and assemble all parts into the final 3D object or scene. Since the backbone is supervised on canonical part decompositions and the slot-gating head is trained on object-level canonical part counts shared across views, the model can infer reasonable canonical structures even under occlusion, while the prototype bank promotes cross-category sharing of recurring structural elements, moving towards a 3D generative world model with structured blocks for diverse real-world objects.

%% file: sec/4_experiment.tex
\section{Experiment}
\label{sec:exp}

\begin{figure*}
  \centering
  \vspace{-5mm}
  \includegraphics[width=\textwidth]{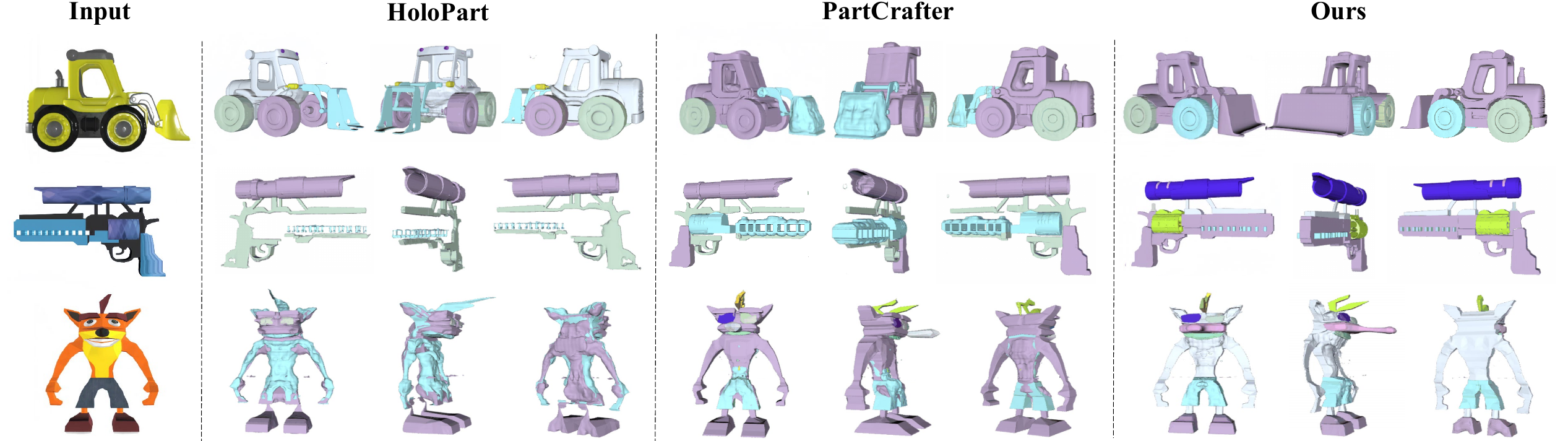}
  \caption{Generalization across diverse 3D object cases. Given an input RGB image (left), 
  our approach reconstructs more complete and structurally coherent shapes with cleaner part boundaries and less part interpenetration.
  }
  \label{fig:result1}
\vspace{-2mm}
\end{figure*}

\subsection{Implementation Details}
\noindent \textbf{Datasets.} Recent advances in large-scale 3D asset aggregation have substantially lowered the barrier to obtaining diverse object geometries, enabling category-agnostic 3D priors to be learned at scale. We train on the high-quality Objaverse \cite{deitke2023objaverse} subset curated by LGM together with ShapeNet-Core \cite{chang2015shapenet} and Amazon Berkeley Objects (ABO) \cite{collins2022abo}. Following PartCrafter \cite{lin2025partcrafter}, we filter training assets with criteria tailored to stable part-level supervision. Specifically, assets without textures are excluded and we further keep only those whose canonical decomposition contains fewer than 16 parts and whose maximum pairwise part overlap, measured by IoU, is below 0.1.  For scene-level generation, we use 3D-Front \cite{fu20213d} for training.

\noindent \textbf{Evaluation Metrics.}
We evaluate structured 3D generation at both the assembled-shape and part levels. Global fidelity is measured by first concatenating all predicted parts into a single mesh and computing the L2 Chamfer Distance and F-Score at threshold 0.1 against the ground-truth mesh. Part independence is quantified by voxelizing each predicted part in canonical space into a $64^3$ grid and averaging the pairwise IoU among parts, where lower IoU indicates less overlap and interpenetration. 

\subsection{Generation Performance Comparison}
\noindent \textbf{Compared Methods.}
We compare against representative single-image 3D baselines: TripoSG \cite{li2025triposg} and HoloPart \cite{yang2025holopart} and PartCrafter \cite{lin2025partcrafter}. For object-composed scene generation on 3D-Front, we additionally compare with MIDI \cite{huang2025midi}, a multi-instance diffusion baseline. 

\input{table/table1}

\input{table/table2}

\begin{figure}
\vspace{-3mm}
\centering
\includegraphics[width=\columnwidth]{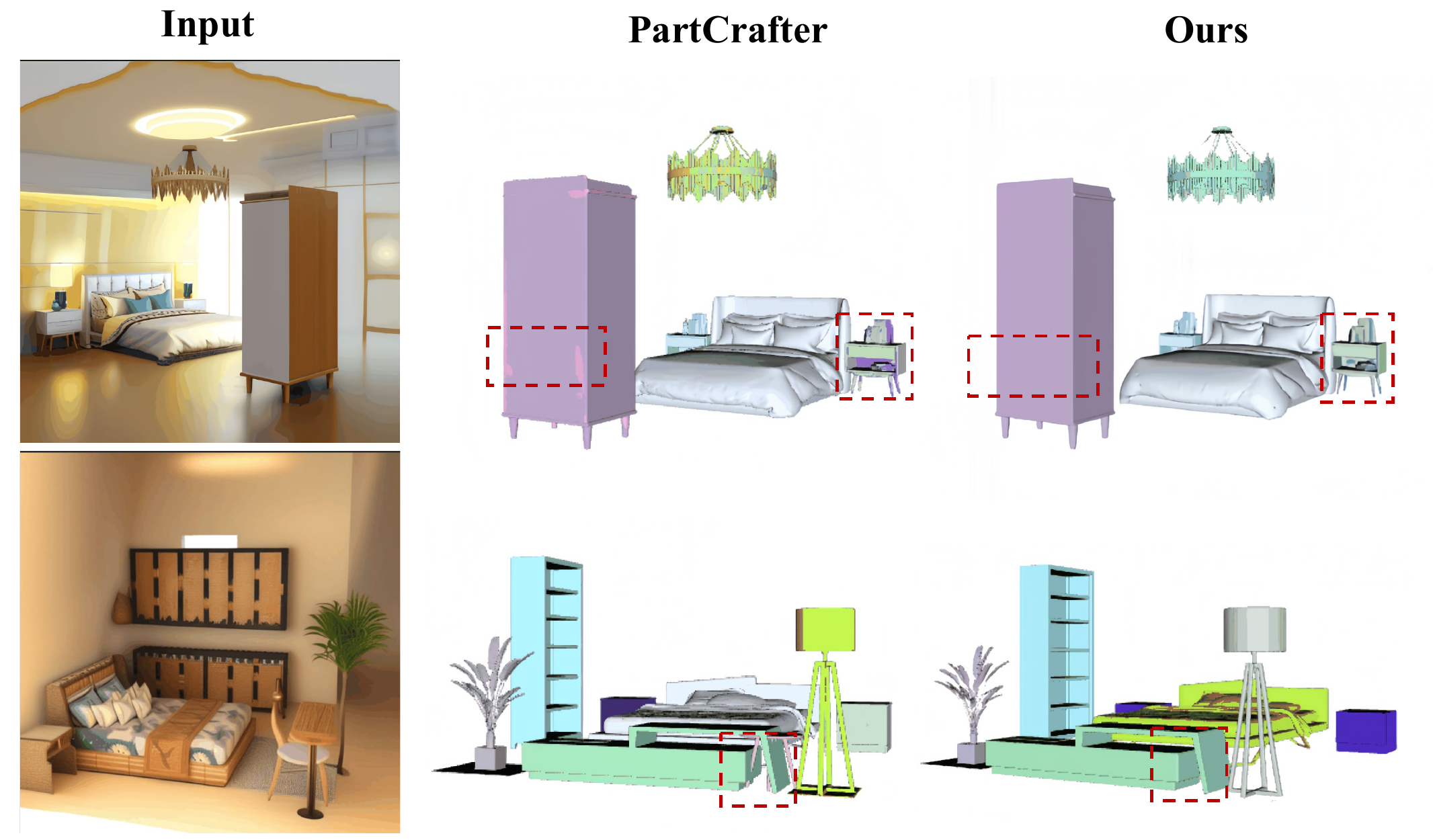}
\caption{Generalization across complicated 3D scene cases.}
\vspace{-2mm}
\label{fig:result2}
\end{figure}

\input{table/table3}

\noindent \textbf{Part-based 3D Object Generation.}
Tab.~\ref{tab:table1} shows that our model consistently improves global fidelity and part independence on both Objaverse and ShapeNet. Compared with monolithic image-to-3D baselines (\textit{e.g.}, TripoSG), our structured generation produces cleaner geometry with substantially better CD/F-Score, while also enabling part-level evaluation. Compared with part-based generators (\textit{e.g.}, HoloPart and PartCrafter), our method achieves lower overlap IoU and higher fidelity, indicating that the generated parts are not only more accurate as a whole but also less interpenetrating and easier to compose. Fig.~\ref{fig:result1} further supports these trends as prior methods may miss thin or articulated components, produce fragmented part boundaries, or introduce noticeable part conflicts, whereas our results maintain more coherent part layouts and preserve fine structures with cleaner separation across different categories.

\noindent \textbf{Compositional 3D Scene Generation.}
We further benchmark object-composed scene generation on 3D-Front. As reported in Tab.~\ref{tab:table2}, our approach improves global fidelity on both the standard split and the occlusion split, with particularly clear gains under occlusion where single-view ambiguity is strongest. Fig.~\ref{fig:result2} highlights typical failure cases of prior compositional generation such as missing or incomplete furniture substructures and local inconsistencies around small objects. In contrast, our model yields more complete object instances and more stable part composition, producing scenes that better match the input while reducing structural artifacts. Overall, these results indicate that learning an adaptive structural hierarchy with shared prototype priors improves both object-level part generation and scene-level composition under challenging single-view conditions.

\subsection{Ablation Study}

Tab.~\ref{tab:ablation} reports a component-wise ablation over three key design choices, including w/o Adaptive Slot Gating (ASG), w/o the Prototype Bank (PB), and w/o the structural warm-up stage. When training is stabilized with warm-up, enabling PB alone already yields strong global fidelity (F-Score $0.751$, CD $0.168$), while enabling ASG alone is less effective (F-Score $0.710$, CD $0.184$), indicating that shared part-level priors are important even before adaptive structure is introduced. Adding ASG on top of PB further improves both assembled-shape accuracy and part separation, reducing overlap (IoU $0.035\!\rightarrow\!0.032$) and improving fidelity (F-Score $0.751\!\rightarrow\!0.781$, CD $0.168\!\rightarrow\!0.159$), showing that image-conditioned slot activation complements prototype-based sharing by better fitting instance-dependent structural complexity. Finally, the warm-up stage is critical for reliable optimization: removing it while keeping ASG and PB causes a sharp degradation (IoU $0.048$, F-Score $0.602$, CD $0.265$), suggesting that directly coupling gating/prototype learning with backbone updates can destabilize denoising and lead to fragmented, overlapping parts. Overall, ASG and PB provide complementary gains, and the staged warm-up is necessary to realize their benefits.

%% file: table/table1.tex
\begin{table*}[t]
    \centering
    \caption{Evaluation on Part-based 3D Object Generation on the Objaverse \cite{deitke2023objaverse}, ShapeNet \cite{chang2015shapenet} and ABO \cite{collins2022abo} datasets.}
    \label{tab:table1}
    \resizebox{\textwidth}{!}{
        \begin{tabular}{l|c|c|ccc|ccc|ccc}
            \toprule
            \multirow{2}{*}{\textbf{Method}} 
            & \multirow{2}{*}{\textbf{\begin{tabular}{@{}c@{}}Segment \\ Mask\end{tabular}}}
            & \multirow{2}{*}{\textbf{\begin{tabular}{@{}c@{}}Fixed Part \\ Number\end{tabular}}}
            & \multicolumn{3}{c|}{\textbf{Objaverse} \cite{deitke2023objaverse}}
            & \multicolumn{3}{c|}{\textbf{ShapeNet} \cite{chang2015shapenet}}
            & \multicolumn{3}{c}{\textbf{ABO} \cite{collins2022abo}} \\
            & & & IoU$\downarrow$ & F-Score$\uparrow$ & CD$\downarrow$
                  & IoU$\downarrow$ & F-Score$\uparrow$ & CD$\downarrow$
                  & IoU$\downarrow$ & F-Score$\uparrow$ & CD$\downarrow$ \\
            \midrule
            Dataset & / & / & 0.080 & / & / & 0.183 & / & / & 0.014 & / & / \\
            \midrule
            TripoSG \cite{li2025triposg} & / & / & / & 0.594 & 0.311 & / & 0.505 & 0.375 & / & 0.710 & 0.202 \\
            HoloPart \cite{yang2025holopart} & $\checkmark$ & $\checkmark$ & 0.044 & 0.692 & 0.192 & 0.111 & 0.550 & 0.351 & 0.045 & 0.809 & 0.134 \\
            PartCrafter \cite{lin2025partcrafter} & $\times$ & $\checkmark$ & 0.036 & 0.747 & 0.173 & 0.029 & 0.567 & 0.321 & 0.024 & 0.862 & 0.105 \\
            \midrule
            Ours & $\times$ & $\times$ & \textbf{0.032} & \textbf{0.781} & \textbf{0.159} & \textbf{0.026} & \textbf{0.573} & \textbf{0.317} & \textbf{0.022} & \textbf{0.891} & \textbf{0.089} \\
            \bottomrule
        \end{tabular}
    }
    \vspace{-4mm}
\end{table*}


%% file: table/table2.tex
\begin{table}
\centering
\vspace{2mm}
\caption{Evaluation on Compositional 3D Scene Generation.}
\resizebox{\columnwidth}{!}{
\label{tab:table2}
\begin{tabular}{l|ccc|ccc}
    \toprule
    \multirow{2}{*}{\textbf{Method}} & \multicolumn{3}{c|}{\textbf{3D-Front} \cite{fu20213d}} & \multicolumn{3}{c}{\textbf{3D-Front (Occluded)} \cite{fu20213d}} \\ 
                    & IoU$\downarrow$ & F-Score$\uparrow$ & CD$\downarrow$ & IoU$\downarrow$ & F-Score$\uparrow$ & CD$\downarrow$ \\ \midrule
    MIDI \cite{huang2025midi}           & \textbf{0.001}          & 0.793        & 0.160          & \textbf{0.002}            & 0.662           & 0.259       \\ 
    PartCrafter \cite{lin2025partcrafter} & 0.003       & 0.815       & 0.149         & 0.004           & 0.780          & 0.151     \\ 
    \midrule
    Ours        & 0.002            & \textbf{0.852}           & \textbf{0.128}         & 0.003            & \textbf{0.785}           & \textbf{0.124}   \\
    \bottomrule
\end{tabular}}
\end{table}

%% file: table/table3.tex
\begin{table}
\caption{Ablation study on the key components of our method over the Objaverse \cite{deitke2023objaverse} dataset.}
\label{tab:ablation}
\centering
\resizebox{0.75\columnwidth}{!}{
\begin{tabular}{@{}c@{}c@{}ccc|ccc@{}}
\toprule
& & \multicolumn{3}{c|}{ {\textbf{Setting}}}  & \multicolumn{3}{c}{\textbf{Metrics}} \\ 
\cmidrule(r){1-5}  \cmidrule(l){6-8}
& & ASG & PB & Warm-up & IoU$\downarrow$ & F-Score$\uparrow$ & CD$\downarrow$\\
\midrule

\multirow{1}{*} &  &  &\checkmark & \checkmark & 0.035  & 0.751 & 0.168 \\
\midrule
\multirow{1}{*} &  & \checkmark & & \checkmark & 0.039 & 0.710 & 0.184 \\
\midrule
\multirow{1}{*} &  & \checkmark & \checkmark &  & 0.048 & 0.602 & 0.265 \\
\midrule
\multirow{10}{*} &  & \checkmark & \checkmark & \checkmark
 & \textbf{0.032} & \textbf{0.781} & \textbf{0.159} \\

\bottomrule
\end{tabular}}
\vspace{-4mm}
\end{table}

%% file: sec/5_conclusion.tex
\section{Conclusion}
\label{sec:con}

We propose a part-to-whole 3D generative model that learns an adaptive structural hierarchy for single-image 3D generation. 
Specifically, it introduces latent structural slots with image-conditioned activation to adapt the part count per instance, and canonical-masked flow supervision to fine-tune stably by supervising only canonical slots. A class-agnostic prototype bank further improves compositional generalization by softly aligning active slots to reusable geometric prototypes and injecting prototype-guided residuals, yielding structured and robust 3D generation under part variation and occlusion.